\documentclass[journal]{IEEEtran}
\ifCLASSINFOpdf
\else
\fi

\usepackage[numbers, sort&compress]{natbib}

\PassOptionsToPackage{table,xcdraw}{xcolor}
\usepackage[english]{babel}
\usepackage{amsmath}
\allowdisplaybreaks
\usepackage{amssymb}
\usepackage{bm}
\usepackage{graphicx}
\usepackage{siunitx}
\usepackage{caption}
\usepackage{subcaption}
\usepackage{multirow}
\usepackage{booktabs}
\usepackage{float}
\usepackage{xspace}
\usepackage[utf8]{inputenc}
\usepackage{doi}

\usepackage{pgfplots}
\usepackage{pgfplotstable}
\pgfplotsset{compat=1.16}
\usepgfplotslibrary{statistics}
\usepgfplotslibrary{colorbrewer}
\usepgfplotslibrary{groupplots}
\usetikzlibrary{matrix}
\usepackage{comment}
\usepackage{dsfont}

\usepackage{array} 
\usepackage{tabularx} 
\renewcommand{\arraystretch}{1.5} 

\DeclareSIUnit{\rad}{rad}
\usepackage{etoolbox}
\newrobustcmd\B{\DeclareFontSeriesDefault[rm]{bf}{b}\bfseries}    

\usepackage{hyperref}

\newcommand{\Transx}[1]{\operatorname{Trans}_{x}(#1)}
\newcommand{\Transy}[1]{\operatorname{Trans}_{y}(#1)}
\newcommand{\Transz}[1]{\operatorname{Trans}_{z}(#1)}
\newcommand{\Rotx}[1]{\operatorname{Rot}_{x}(#1)}
\newcommand{\Roty}[1]{\operatorname{Rot}_{y}(#1)}
\newcommand{\Rotz}[1]{\operatorname{Rot}_{z}(#1)}

\begin{document}
%
\title{A Unified Calibration Framework for High-Accuracy Articulated Robot Kinematics
}

\author{
Philip Tobuschat,
Simon Duenser,
Markus Bambach,
Ivo Aschwanden%
\thanks{All authors are with the Advanced Manufacturing Laboratory, Department of Mechanical and Process Engineering, ETH Zurich, 8005 Zurich, Switzerland. Corresponding author: Philip Tobuschat (\texttt{philipt@ethz.ch}).}%
\thanks{This work has been submitted to the IEEE for possible publication. Copyright may be transferred without notice, after which this version may no longer be accessible.}%
}

\maketitle


\begin{abstract}
Researchers have identified various sources of tool positioning errors for articulated industrial robots and have proposed dedicated compensation strategies. However, these typically require individual, specialized experiments with separate models and identification procedures. This article presents a unified approach to the static calibration of industrial robots that identifies a robot model, including geometric and non-geometric effects (compliant bending, thermal deformation, gear transmission errors), using only a single, straightforward experiment for data collection. The model augments the kinematic chain with virtual joints for each modeled effect and realizes the identification using Gauss-Newton optimization with analytic gradients. Fisher information spectra show that the estimation is well-conditioned and the parameterization near-minimal, whereas systematic temporal cross-validation and model ablations demonstrate robustness of the model identification. The resulting model is very accurate and its identification robust, achieving a mean position error of 26.8 $\si{\micro \metre}$ on a KUKA KR30 industrial robot compared to 102.3 $\si{\micro \metre}$ for purely geometric calibration.
\end{abstract}



%
\IEEEpeerreviewmaketitle

\section{Introduction}
\label{sc:Introduction}
\subsection{Background}
\label{sc:Background}
Articulated industrial robots are known for their great repeatability, enabling their usage for applications with repetitive movements such as pick-and-place, soldering, welding, spray painting, and medical applications. Many other applications, such as machining applications \cite{zhu_robotic_2020}, have so far been mostly out of reach for articulated robots due to the requirement of high absolute accuracy \cite{kim_robotic_2019}. Enabling the widespread use of articulated robots for such tasks would significantly reduce manufacturing costs, which makes the investigation of further methods for accuracy improvement critical.
To improve accuracy without making hardware changes, it is standard practice to use a calibrated robot model so that manufacturing and assembly errors and other unwanted effects may, while not eliminated, be compensated for. The main known sources of error are the elastic deformation under load, referred to as compliance \cite{abele_modeling_2007}, thermal deformations \cite{heisel_thermal_1997}, and gear transmission errors that manifest as non-linear joint behavior \cite{garcia_compact_2020}.
While various modeling approaches for different individual effects have been proposed, a comprehensive framework for combined calibration remains absent. Existing methods that address more than one effect often either rely on strong modeling assumptions \cite{whitney_industrial_1986, nubiola_absolute_2013}, which limits the accuracy of the resulting model, or identify effects sequentially \cite{gong_nongeometric_2000}, which may result in suboptimal trade-offs between effects in the identification process, thus also limiting accuracy. 
Therefore, a unified identification process would be necessary to achieve highly accurate robot calibration. Furthermore, independently of the identification procedure, no model has been suggested that explicitly accounts for the \emph{combined} effects of all the listed effects. However, as our results will show, none of these can be neglected when high accuracy is the goal. 
Building a unified framework poses several interconnected challenges. Firstly, it necessitates a dataset that contains all the appropriate information. This may be challenging because compliance errors change on a macro-scale of joint states and couple the states in a nonlinear manner, gear transmission errors change on a micro-scale of joint states, and thermal effects are subject to the external change of the ambient temperature. 
Secondly, the co-identification of all of these effects introduces numerical challenges that need to be handled with care for robust results.
The calibration framework we propose aims to address these issues.
We show experimentally that it is possible to effectively capture geometry, compliance, joint behavior, and environmental thermal effects from a set of randomly sampled data points, given that the set is large enough and contains sufficient variations in all inputs.
Our model consists of four submodels, each targeting calibration for a specific class of effects:
\vspace{-0.5em}
\begin{itemize}
  \setlength\itemsep{0em}
  \item \textbf{Geometry:} Accounts for deviations from nominal link dimensions.
  \item \textbf{Compliance:} Models each joint as a one-dimensional spring to capture elastic deformation in the robot.
  \item \textbf{Thermal:} Captures temperature-induced link expansion due to environmental conditions.
  \item \textbf{Joint correction:} Maps commanded joint angles to effective joint angles to compensate for gear transmission errors and other joint-related inaccuracies.
\end{itemize}
\vspace{-0.5em}
To make the calibration process computationally feasible, we introduce models for all effects that are differentiable with respect to calibration parameters. This enables the use of efficient, gradient-based optimization. We investigate the conditioning of the identification for all considered effects and support the robustness of our method with additional numerical tests.
\subsection{Related Work}
\label{sc:RelatedWork}
Several mathematical frameworks have been suggested to model the forward kinematics of articulated robots and their geometric calibration. The Denavit-Hartenberg (DH) model \cite{denavit_kinematic_1955} is well-established and uses interpretable robot dimensions as parameters that may be re-identified for the purpose of calibration. Various expansions and generalizations of the DH model, such as the Stone model, also called S-model, \cite{stone_kinematic_1987}, or the complete and parametrically continuous model in \cite{zhuang_complete_1990} have been suggested. In line with these models, we formulate our forward kinematic chain using homogeneous transformations. Calibration to non-geometric errors, on the other hand, is not included in these models, making more advanced frameworks necessary.
Compliance is often the first non-geometric effect tackled as it can describe a significant portion of end-effector errors but requires a relatively small amount of parameters and data for identification. \cite{whitney_industrial_1986} estimates the effective stiffness of two of the joints experimentally. \cite{gong_nongeometric_2000} introduces a compliance model for all joints under end-effector load while making additional assumptions for the self-gravity case. This enables the compensation of a significant fraction of compliance errors using only three parameters. \cite{nubiola_absolute_2013} presents a further expanded model that includes compliance in four out of six joints and uses load under self-weight from the mass of the links as provided by the manufacturer. Other comprehensive approaches to compliance identification are found for example in \cite{chen_stiffness_2019, dumas_joint_2011} with a recent survey of used approaches presented by \cite{zhu_high_2022}. While there is a clear trend towards more complex compliance models, what the suggested approaches still have in common, is that they avoid the combined identification of joint stiffness values and link masses from static measurements by the use of strong modeling assumptions, or, in the case of \cite{nubiola_absolute_2013}, the use of prior knowledge. This is a shortcoming that our method addresses.
Another important effect is imperfections in the gear and transmission system, which manifests in the form of nonlinear, erroneous joint behavior. We refer to these collectively as joint errors. The magnitude and shape of such errors greatly depend on the type of transmission principle used; \cite{garcia_compact_2020} provides a comprehensive summary.
Harmonic drives are generally considered one of the more precise types of gear systems in broad use, but are still subject to errors that have been experimentally reported to be in the range of up to 1 arcmin \cite{garcia_compact_2020, nye_ted_w_kraml_robert_p_harmonic_1991, jia_modeling_2021, m_iwasaki_modeling_2009, sammons_modeling_2014, slamani_characterization_2013}. \cite{r_judd_technique_1987} first used a Fourier series to compensate for gear transmission errors, leveraging their largely oscillatory nature. \cite{m_iwasaki_modeling_2009} expand on this idea by including those sinusoidal basis functions corresponding to specific harmonic orders of an identified base frequency.
\cite{ma_modeling_2018} demonstrate a compensation mechanism for gear errors in all joints, using a basis of Chebyshev polynomials, also leveraging their periodicity properties. 
These previously studied modeling approaches emphasize data efficiency via the use of a small number of periodic basis functions to derive a rough compensation mechanism. While we also observe a joint behavior with clearly visible oscillatory modes, we aim for more accurate modeling by dropping the enforcement of periodicity and using a fine-grained piece-wise linear function.
Research on thermal effects in articulated robots is still relatively scarce. 
\cite{gong_nongeometric_2000} identify a model for geometric and compliance errors at room temperature, and then freeze those parameters to identify thermal effects using specifically designed experiments where the robot is heated and cooled down. A regression model is used to predict thermal errors. Their approach effectively enables the compensation for thermal errors for the analyzed movement patterns, but lacks thorough validation and does not consider the combined optimality of the individual identifications.
\cite{poonyapak_towards_2006} compensate for thermal drift at the end-effector using a linear, physics-inspired model. Their approach is limited by the size of the dataset for identification and validation. 
\cite{sigron_compensation_2024} build a similar linear thermal compensation model and paid special attention to different thermal states in the training and validation set.
\cite{li_dynamic_2016} employs finite element theory to build a temperature and deformation model, and to aid the compensation of thermal effects, which is realized as a linear regression between measured temperature and changes in geometric parameters.
Collectively, these studies experimentally support the applicability of linear compensation approaches for thermal errors. In terms of data collection and model validation, we observe varying approaches. The integration of thermal compensation mechanisms into broader calibration schemes has seen little investigation and a fully combined, simultaneous identification has not been suggested.
Although models for geometry, compliance, joint behavior, and thermal effects have each been studied in isolation, no prior work has tackled their simultaneous identification in a unified calibration. As a result, fully combined parameter co-estimation across all these non-geometric effects remains an open challenge.
\subsection{Contribution}
\label{sc:Contribution}
The core contribution of our work is the presentation of a unified calibration framework for the most prominent adverse effects in industrial articulated robots. 
We thoroughly investigate several important characteristics of the suggested method that one would demand from a practically useful calibration method. 
Our primary objective is the highest possible accuracy.
We evaluate the achieved accuracy experimentally using a rigorous cross-validation scheme. To help the interpretation of these results, we additionally determine our setup's repeatability experimentally.
Secondly, the calibration should be robust. We experimentally verify several key aspects that play into a calibration's robustness: the numerical conditioning of the identification procedure, the variation of the estimated parameters between training sets, and the accuracy and generalization behavior under varying amounts of training data. We also demonstrate robustness to the initialization by using an uninformed initial guess. 
Finally, we emphasize the practical applicability. Our framework is designed to require no prior knowledge about a specific robot and only a single, simple experiment; making the overall calibration procedure straightforward.
\subsection{Structure}
\label{sc:Structure}
This article is structured as follows. In Sec.~\ref{sc:RobotModel} we introduce the full robot model. Sec.~\ref{sc:Identification} introduces the identification problem and the optimization algorithm we employ to solve it. Sec.~\ref{sc:ExpSetup} introduces the experimental setup and the data collection process. Sec.~\ref{sc:ModelValidation} presents the experiments we perform to validate the identification. Sec.~\ref{sc:Repeatability} presents the experimentally determined repeatability of our setup. Sec.~\ref{sc:Results} presents the main results, which we discuss in Sec.~\ref{sc:Discussion}. Sec.~\ref{sc:Conclusion} presents the conclusion.
\section{Robot Model}
\label{sc:RobotModel}
\subsection{Base Robot Model}
\label{sc:UncalibratedRobotModel}
We use a forward-kinematics model based on homogeneous transformations, $\bm{T} \in \mathrm{SE}(3)$, such that
\begin{equation}
    \bm{T} = \begin{pmatrix} \bm{R} & \bm{t} \\ \bm{0} & 1 \end{pmatrix},
\end{equation}
where $\bm{R} \in \mathrm{SO}(3)$ corresponds to a rotation and $\bm{t} \in \mathbb{R}^3$ to a translation. The transformation of the tool-center-point (TCP) frame into the global frame can then be written as
\begin{equation}
    \label{eq:RobotModelUncalibrated}
    \bm{T}^{\mathrm{TCP}}_{\mathrm{glob}}(\bm{q}) = \bm{T}^{\mathrm{base}} \bm{T}_1^{\mathrm{joint}}(q_1) \bm{T}_1^{\mathrm{link}} \cdots \bm{T}_6^{\mathrm{joint}}(q_6) \bm{T}_6^{\mathrm{link}} \bm{T}^{\mathrm{TCP}}_{\mathrm{local}}.
\end{equation}
The joint transformations depend on the robot's joint angles, summarized in the vector $\bm{q} = (q_1, \dots, q_6)^{\mathrm{T}}$. Without loss of generality, we assume joint rotations around principal axes. The link transformations contain only the links' nominal dimensions as translation vectors, $\bm{t}_1, \dots, \bm{t}_6$. This yields
\begin{equation}
    \label{eq:JointTransformation}
    \bm{T}_i^{\mathrm{joint}}(q_i) = \begin{pmatrix} \bm{R}_i^{\mathrm{joint}}(q_i) & \bm{0} \\ \bm{0} & 1 \end{pmatrix} \quad \text{and} \quad \bm{T}_i^{\mathrm{link}} = \begin{pmatrix} \bm{0} & \bm{t}_i \\ \bm{0} & 1 \end{pmatrix}.
\end{equation}
\subsection{Augmented Robot Model}
\label{sc:CalibratedRobotModel}
\begin{figure}[!ht]
    \centering
    \includegraphics[scale=1.1]{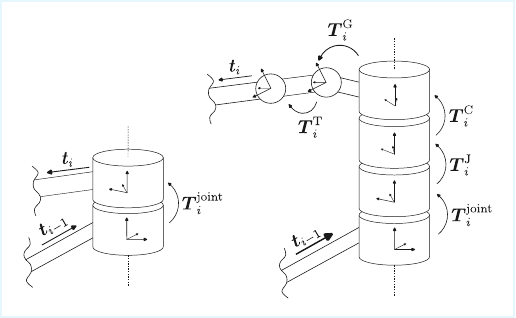}
    \caption{Visualization of the joint modeling: unaugmented (left) versus augmented with virtual joints (right).}
    \label{fig:JointModel}
\end{figure}

Our approach to calibration consists of introducing \emph{virtual joints} individually for each of the effects we want to compensate for, per real joint. Similar to a real joint, a virtual joint introduces a transformation into the kinematic chain according to some underlying model. For us this includes geometry ($\bm{T}^{\mathrm{G}}$), compliance ($\bm{T}^{\mathrm{C}}$), nonlinear joint correction ($\bm{T}^{\mathrm{J}}$), and thermal effects ($\bm{T}^{\mathrm{T}}$). As a result, the transformation of each real joint is expanded according to
\begin{equation}
    \label{eq:virtualjoints}
    \bm{T}_i^{\mathrm{joint}} \rightarrow \bm{T}_i^{\mathrm{joint}}
    \bm{T}_i^{\mathrm{J}} \bm{T}_i^{\mathrm{C}}
    \bm{T}_i^{\mathrm{G}} \bm{T}_i^{\mathrm{T}},
\end{equation}
which is visualized in Figure~\ref{fig:JointModel}.
Each virtual joint transformation is in its most general form parameterized by three Euler angles $\zeta$, $\xi$, and $\chi$ as well as the translational components $x$, $y$, and $z$. We write collectively: $\bm{p} = [\zeta, \xi, \chi, x, y, z]^{\mathrm{T}}$, so that for each virtual joint $\bm{T} = \bm{T}(\bm{p})$, whereby $\bm{p}$ in turn captures dependencies on the vector $\bm{\theta}$ of calibration parameters. Subscripts specify the joint and type of correction. Including the augmentation, we now introduce the full robot model as a function of the joint states $\bm{q}$ and the ambient temperature $\kappa$, and parameterized by $\bm{\theta}$. We include explicit functional dependencies of each $\bm{p}$ for clarity, which we expand on in Sec.~\ref{sc:VirtualJointModeling}, so that the full, augmented robot model reads
\begin{equation}
\label{eq:FullRobotModel}
\begin{aligned}
    \bm{T}^{\mathrm{TCP}}_{\mathrm{glob}} \left( \bm{q}, \kappa ; \bm{\theta} \right) =
    & \bm{T}^{\mathrm{base}} 
    \bm{T}_0^{\mathrm{G}}\left(\bm{p}_{\mathrm{G}, 0}\right)  \\
    & \prod_{i=1}^{6} \bigg(
    \bm{T}_i^{\mathrm{joint}}\big(q_i\big) \;
    \bm{T}_i^{\mathrm{J}}\big(\bm{p}_{\mathrm{J}, i} (q_i) \big) \;
    \bm{T}_i^{\mathrm{C}}\big(\bm{p}_{\mathrm{C}, i} (\bm{q}) \big) \\
    & \bm{T}_i^{\mathrm{G}}\big(\bm{p}_{\mathrm{G}, i}\big) \;
    \bm{T}_i^{\mathrm{T}}\big(\bm{p}_{\mathrm{T}, i} (\kappa) \big) \;
    \bm{T}_i^{\mathrm{link}} \bigg) \bm{T}^{\mathrm{TCP}}_{\mathrm{local}}.
\end{aligned}
\end{equation}
Apart from the introduced joint expansion,~\eqref{eq:FullRobotModel} additionally introduces a geometric transformation, $\bm{T}_0^{\mathrm{G}}$, to localize the robot's base.
\subsection{Submodel Parametrization}
\label{sc:VirtualJointModeling}
In the following, we introduce our four submodels that combine into our full suggested model: geometry, compliance, thermal, joint correction. Specifically, this encompasses the calculation of the transformation parameters $\bm{p}$ for each submodel. We will additionally introduce the derivatives of these with respect to calibration parameters, which will be required for the optimization process.
\paragraph{Compliance Joints} Compliance under load is typically modeled by lumping the deformation of a joint and the following link into a single virtual spring in the joint, this is known as the elastostatic model. While for dynamic settings, the spring behavior is sometimes assumed in several or even all six dimensions, for the static case, a one-dimensional, rotational spring model for every joint is usually sufficient~\cite{g_alici_enhanced_2005, slavkovic_method_2014, dumas_joint_2011, abele_modeling_2007}. Some researchers have suggested further model reduction and have included compliant behavior only in a few selected joints~\cite{nubiola_absolute_2013, gong_nongeometric_2000}. While such a reduced model may exhibit improved robustness and data-efficiency of the identification process, due to the smaller number of parameters, most authors agree that for high accuracy, every joint’s compliance should be included. We therefore do not perform any such reductions and stick to the established approach that assumes compliant behavior in every joint.
This is also in line with our goal of avoiding custom modeling for any specific robot design.
Consequently, the parameter vector for compliance joints, $\bm{p}_{\mathrm{C}, i}$, is zero for all elements except one rotational component. 
We introduce a one-hot selection vector $\bm{s}_i \in \{0,1\}^6$ that indicates the active rotational direction for each joint. 
This allows us to write
\begin{equation}
\label{eq:ComplianceSpring}
 \bm{p}_{\mathrm{C}, i} = p^\star_{\mathrm{C}, i} \bm{s}_i, \quad \text{with} \quad  p^\star_{\mathrm{C}, i} = k_i \tau^\star_i,
\end{equation}
where $k$ represents the virtual spring's compliance, and $\tau^\star_i$ is the torque in the direction of the joint's rotation. The loads we consider are the robot's weight and the tool's weight. This load is modeled by attributing a single mass, $m_i$, to each link, which we include as calibration parameters. The mass of the links, the transmission system, the motors, and the tool is thus lumped into these mass parameters. The center of mass is assumed to be located along the link vectors $\bm{t}_i$ at a ratio of $r_i$ of the link. For a given pose $\bm{q}$, the torque on joint $i$ is then computed as follows, whereby all vectors are assumed to be represented in the global coordinate frame for ease of notation:
\begin{equation}
    \bm{\tau}_i = r_i m_i \bm{t}_i \times \bm{g} + \sum_{j = i + 1}^{6} m_j \left( \bm{t}_{i \rightarrow j} + r_j \bm{t}_j \right) \times \bm{g}.
    \label{eq:LoadCalculation}
\end{equation}
Here, $\bm{t}_i$ is the translational component of the link transformation introduced in~\eqref{eq:RobotModelUncalibrated}, $\bm{g} = [0, 0, 9.81 \si{\metre \per \second \squared}]^\mathrm{T}$ is the gravitational acceleration, and $\bm{t}_{i \rightarrow j}$ is the translation vector from joint $i$ to $j$ of the uncalibrated robot model, that is found as the translational component of the subchain $\bm{T}_i^{\mathrm{joint}}(q_i) T_i^{\mathrm{link}} \dots T_{j-1}^{\mathrm{link}}$ of the uncalibrated forward kinematics from~\eqref{eq:RobotModelUncalibrated}. The torque acting on the \emph{spring} $\tau^\star_i$ is then calculated from $\bm{\tau}_i$ by projecting into the joint's local frame and isolating the relevant component using $\bm{s}_i$.
The mass and compliance parameters are combined into 
\begin{equation}
    \bm{m} = [m_1, \dots, m_6]^T, \; \bm{k} = [k_1, \dots, k_6]^T, \; \text{and} \; \bm{\theta}_\mathrm{C} =\begin{bmatrix} \bm{m} \\ \bm{k} \end{bmatrix}.
\end{equation}
both of which are identified as calibration parameters. For simplicity, we assume the centers of mass at the midpoint of each link, so that $r_i = \tfrac12$ for all joints $i=1,\dots,6$. We thereby avoid relying on manufacturer-provided information. Even though the assumed values for $r_i$ might not necessarily be accurate, we will see that overall accuracy can nonetheless be achieved through the flexibility of identifying mass and compliance parameters. While it may appear natural to include all of $r_i$ into the vector of calibration parameters, we found that doing so leads to poor conditioning of the problem.
Note that the suggested compliance model can be identified from data only up to one remaining degree of freedom. As can be seen from~\eqref{eq:ComplianceSpring} and~\eqref{eq:LoadCalculation}, multiplying the masses $\bm{m}$ by a constant $c \in \mathbb{R}_{\ne 0}$ and dividing the compliances $\bm{k}$ by that value will result in the same transformations for all virtual joints. We reduce this set of solutions to a single one by fixing one of the values during the optimization process. For this, we choose the mass $m_6$, which primarily captures the tool weight.
Additionally, since we only consider gravitational load, joint 1 experiences no load in the direction of the joint's axis, as would be the case for any joint whose axis is always parallel to the gravity vector. For our setting, this makes $m_1$ and $k_1$ unidentifiable. 
Differentiating~\eqref{eq:LoadCalculation} with respect to mass $m_j$ yields
\begin{equation}
    \frac{\partial \bm{\tau}_i}{\partial m_j} = 
    \begin{cases}
        {0}, & \text{if } j < i, \\
        r_j \bm{t}_j \times \bm{g}, & \text{if } j = i, \\
        \left( \bm{t}_{i \rightarrow j} + r_j \bm{t}_j \right) \times \bm{g}, & \text{if } j > i.
    \end{cases}
    \label{eq:LoadCalculationDerivative}
\end{equation}
This can be used for the necessary derivatives of~\eqref{eq:ComplianceSpring} with respect to calibration parameters, so that
\begin{equation}
    \label{eq:ComplDiff1}
    \frac{\partial p_{\mathrm{C}, i}^\star}{\partial m_j} = k_i \frac{\partial \tau^\star_i}{\partial m_j}
\end{equation}
and
\begin{equation}
    \label{eq:ComplDiff2}
    \frac{\partial p_{\mathrm{C}, i}^\star}{\partial k_j} = 
    \begin{cases}
        \tau^\star_i, & \text{if } j = i, \\
        {0}, & \text{else }
    \end{cases}
\end{equation}
for any combination of joint $i$ and parameters with index $j$.
\paragraph{Thermal Joints} Our thermal model introduces a linear relation between link elongation and temperature. This is in line with the established physical relation for homogeneous materials and small temperature excursions, and has further been confirmed in experimental studies to be a reasonable approximation for thermal models for industrial robots~\cite{sigron_compensation_2024, li_dynamic_2016}. The parametrization of the thermal joints can thus be written as
\begin{equation}
\label{eq:ThermalCompensationModel}
\bm{p}_{\mathrm{T}, i} = \begin{bmatrix} \bm{0} \\ \alpha_i (\kappa - \kappa_0) \bm{t}_i  \end{bmatrix}\quad \text{and} \quad \frac{\partial \bm{p}_{\mathrm{T}, i}}{\partial \alpha_i} = \begin{bmatrix} \bm{0} \\  (\kappa - \kappa_0) \bm{t}_i \end{bmatrix}.
\end{equation}
 Here, $\kappa$ is the current ambient temperature measurement, and $\kappa_0$ is the temperature baseline, which we set to 25\si{\celsius}. The link translations, $\bm{t}_i$, have been introduced in~\eqref{eq:JointTransformation}. The coefficient of thermal expansion for each link is introduced as $\alpha_i$.
The thermal calibration parameters are combined via
\begin{equation}
    \bm{\theta}_\mathrm{T} = [ \alpha_1, \dots, \alpha_6 ]^T.
\end{equation}
\paragraph{Geometric Joints} Geometric calibration aims to account for offsets between nominal and real link dimensions due to manufacturing imprecision. We additionally include the localization of the robot base and the determination of tool dimension into this category.
We introduce the geometric calibration in the form of a transformation with five degrees of freedom. These are the Euler angles, and the two translational degrees of freedom that are orthogonal to the joint axis. 
As a result, only the base transformation needs to contain all six calibration parameters, whereas all other geometric transformations, each following a real joint, have one translational element that is not identified and kept at zero to avoid over-parametrization. The first joint, its rotation around the local z-axis, is thus parametrized as
\begin{align}
    \bm{p}_{\mathrm{G}, 1} =& [\zeta_{\mathrm{G}, 1}, \xi_{\mathrm{G}, 1}, \chi_{\mathrm{G}, 1}, x_{\mathrm{G}, 1}, y_{\mathrm{G}, 1}, 0]^T, \\
    \bm{\theta}_{\mathrm{G}, 1} =& [\zeta_{\mathrm{G}, 1}, \xi_{\mathrm{G}, 1}, \chi_{\mathrm{G}, 1}, x_{\mathrm{G}, 1}, y_{\mathrm{G}, 1}]^T, \text{ and thus } \\
    \frac{\mathrm{d} \bm{p}_{\mathrm{G}, 1}}{\mathrm{d} \bm{\theta}_{\mathrm{G}, 1}} =& \mathbb{I}_{6 \times 5}.
\end{align}
The remaining joints are handled analogously based on each joint's rotation axis, with all parameters being collected via
\begin{equation}
    \bm{\theta}_\mathrm{G} = [\bm{\theta}_{\mathrm{G}, 0}^T, \dots, \bm{\theta}_{\mathrm{G}, 6}^T]^T.
\end{equation}
The translation along the joint axis is excluded, since its effect is \emph{independent} of the joint rotation and therefore can be equivalently captured by the translation of the previous virtual joint. Thus, only the robot base requires full parametrization with six degrees of freedom. We'll derive this result formally below. It is worth noting that for a model with purely kinematic compensation, a reduction to four degrees of freedom per joint would be possible, as is the standard, e.g. for DH-based geometric calibration methods. However, in our case, this further reduction is not possible due to our thermal compensation. To show this, we consider the following setting.
Let $T_1$, $T_2$ $\in$ $\mathrm{SE(3)}$ be fixed transformations associated with the geometric calibration of the robot model such that the subchain $\tilde{T} = T_1 \; T^{\mathrm{joint}}(\tilde{q}) \; T_2$ for some joint transformation $T^{\mathrm{joint}}(\tilde{q}) = \mathrm{Rot}_z(\tilde{q})$ occurs in the kinematic chain. 
Without loss of generality, we assume a coordinate frame such that the joint's axis is aligned with the local z-axis. 
Recall that any general transformation can be written as a concatenation of translation and rotation primitives in arbitrary order for an appropriate choice of parameters $p = [\phi_x, \phi_y, \phi_z, d_x, d_y, d_z]^T$, allowing us to split $T_1$ and $T_2$ into parts parallel and perpendicular to the joint's assumed z-axis so that
\begin{equation}
    T_1 = T_{1\perp z} \; T_{1\parallel z} \quad \text{and} \quad T_2 = T_{2\parallel z} \; T_{2\perp z}
\end{equation}
with
\begin{equation}
\begin{aligned}
T_{\tilde{i}\perp z}&=\Transx{d_{\tilde{i}x}}\Rotx{\phi_{\tilde{i}x}}\Transy{d_{1y}}\Roty{\phi_{\tilde{i}y}}, \; \text{and }\\[4pt] \
T_{\tilde{i}\parallel z}&=\Transz{d_{\tilde{i}z}}\Rotz{\phi_{\tilde{i}z}}, \quad \text{for} \quad \tilde{i} = 1, 2.\\[4pt]
\end{aligned}
\end{equation}
We now introduce the thermal transformation $T_\mathrm{T}(\tilde{\kappa})$ as a function of the temperature $\tilde{\kappa}$, which, following~\eqref{eq:ThermalCompensationModel}, contains only a translation. This allows us to write the considered subchain as follows:
\begin{align}
\tilde{T} &\stackrel{(1)}{=} T_1\,T_{\!T}(\kappa) \; T^{\text{joint}}(\tilde{q}) \; T_2 \notag\\[2pt]
&\stackrel{(2)}{=}
  T_{1\perp z} T_{1\parallel z}\,
 T_\mathrm{T}(\kappa) \;
\Rotz{\tilde{q}} \; T_{2\parallel z} \; T_{2\perp z}
\notag\\[2pt]
&\stackrel{(3)}{=}
  T_{1\perp z} T_{1\parallel z}\,
 T_\mathrm{T}(\kappa) \; T_{2\parallel z} \; 
\Rotz{\tilde{q}} \; T_{2\perp z}
\notag\\[2pt]
&\stackrel{(4)}{=}
  T_{1\perp z} \Rotz{\phi_{1z}} \Transz{d_{1z}} \,
 T_\mathrm{T}(\kappa) \notag\\
& \quad \; \Transz{d_{2z}}\Rotz{\phi_{2z}} \; 
\Rotz{\tilde{q}} \; T_{2\perp z}
\notag\\[2pt]
&\stackrel{(5)}{=}
  T_{1\perp z} \Rotz{\phi_{1z}} \Transz{d_{1z} + d_{2z}} \, \notag\\
& \quad T_\mathrm{T}(\kappa) \; \Rotz{\phi_{2z}} \; 
\Rotz{\tilde{q}} \; T_{2\perp z}
\notag\\[2pt]
&\stackrel{(6)}{=}
  T_{1\perp z} \Rotz{\phi_{1z}} \Transz{d_{z}'} \, \notag\\
& \quad T_\mathrm{T}(\kappa) \; \Rotz{\phi_{2z}} \; 
\Rotz{\tilde{q}} \; T_{2\perp z}.
\end{align}
Equalities $(3)$ and $(4)$ hold with similar reasoning as above, whereas equality $(5)$ holds due to the commutative property between translations. We see that we were able to reduce one pair of parameters $d_{1z}, d_{2z}$ into the combined parameter $d_{z}'=d_{1z}+d_{2z}$, leaving the transformation $T_2$ defined through five free parameters.
Note that when no thermal transformation is present, a further reduction may be made:
\begin{equation}
\tilde{T}
 \stackrel{(7)}{=} 
 T_{1\perp z}\,
    \Transz{\,d_{z}'}\,
    \Rotz{\phi_{z}'}\,
    \Rotz{\tilde{q}}\,
    T_{2\perp z},
\end{equation}
Where we used $\phi_{z}'=\phi_{1z}+\phi_{2z}$, allowing us to retrieve a four-DoF-per-joint parameterization similar to the DH model.
\paragraph{Joint Correction} Errors in the gear transmission system can result in a non-identity relation between commanded and real joint angles. We introduce a corrective rotation right after the real joint's rotation to account for the offset between the two. Therefore, we will be left with a single non-zero component, $p_{\mathrm{J}, i}^\star$, of the parameter vector in the direction of the joint's rotation to capture this offset. We use a piece-wise linear function that maps from the commanded joint angle, $q_i$, to $p_{\mathrm{J}, i}^\star$. We let the $n_\mathrm{supp}$ support points of this function be evenly spaced between the minimal and maximal joint angles present in the training data, $q_{i, \mathrm{min}}$ and $q_{i, \mathrm{max}}$, and let the calibration parameters $\bm{\theta}_{\mathrm{J}, i} = [ \theta_{\mathrm{J},i,1}, \dots, \theta_{\mathrm{J},i,{n_\mathrm{supp}}} ]^\mathrm{T} $ be the function values for each support point. For all joints, we introduce the same density of support points $d_\mathrm{supp}$, so that the spacing between support points is \( q_{i, \mathrm{step}} = \frac{1}{d_{\mathrm{supp}}} \). For the unique pair of support points $j_{-}$ and $j_{+}$ for which a joint angle $q_i$ lies between $j_{-} \cdot q_{i, \mathrm{step}}$ and $j_{+} \cdot q_{i, \mathrm{step}}$, we can write the piece-wise linearity as an interpolation between the two relevant values $\theta_{\mathrm{J},i,j_{-}}$ and $\theta_{\mathrm{J},i,j_{+}}$ of the form
\begin{equation}
    \label{eq:JointCorrection}
    p_{\mathrm{J}, i}^\star = \theta_{\mathrm{J},i,j_{-}} + \frac{q_{i} - j_{-} \cdot q_{i, \mathrm{step}}}{q_{i, \mathrm{step}}} ( \theta_{\mathrm{J},i,j_{+}} -\theta_{\mathrm{J},i,j_{-}} ).
\end{equation}
The derivative of~\eqref{eq:JointCorrection} with respect to calibration parameters for a given joint angle then contains only two non-zero values:
\begin{equation}
\frac{\partial p_{\mathrm{J}, i}^\star}{\partial \bm{\theta}_{\mathrm{J}, i}} = 
\begin{bmatrix}
0, \dots,0, 1 - \frac{q_i - j_{-} \cdot q_{i, \mathrm{step}}}{q_{i, \mathrm{step}}}, \frac{q_i - j_{-} \cdot q_{i, \mathrm{step}}}{q_{i, \mathrm{step}}},0,\dots, 0
\end{bmatrix}^\mathrm{T}.
\end{equation}
The joint correction calibration parameters are combined via
\begin{equation}
    \bm{\theta}_\mathrm{J} = [\bm{\theta}_{\mathrm{J}, 1}^T, \dots, \bm{\theta}_{\mathrm{J}, 6}^T]^T.
\end{equation}
We will refer to the collective modeling choices introduced here as our joint model.
\noindent
\paragraph{Combined Parameters} The parameters of all virtual joints combine into the full parameter vector
\begin{equation}
    \label{eq:Theta}
    \begin{split}
        \bm{\theta} = [ \bm{\theta}_\mathrm{G}^T, \bm{\theta}_\mathrm{C}^T, \bm{\theta}_\mathrm{T}^T,  \bm{\theta}_\mathrm{J}^T ]^T.
    \end{split}
\end{equation}
Note that this parameter vector $\bm{\theta}$ corresponds to our \textbf{full model}. In Sec.~\ref{sc:Results} we will present results for this model as well as for reduced versions, whereby the capital letters will indicate, which submodels are used, i.e. which types of parameters are included in the parameter vector. As an example, the \textbf{GC model} includes only geometry and compliance, so that the parameter vector would become $
\tilde{\bm{\theta}} = \Big[ \bm{\theta}_{\mathrm{G}}^T,  \bm{\theta}_{\mathrm{C}}^T\Big]^T$. 
\noindent
The initial guess, $\bm{\theta}^{(0)}$, for the combined parameter vector rev{is set to ones for the masses and zeros for all remaining parameters}. As can be seen from~\eqref{eq:ComplDiff1} and ~\eqref{eq:ComplDiff2}, initializing exclusively with zeros would result in a saddle point. We write
\begin{equation}
\label{eq:InitialGuess}
\bm{\theta}_{\mathrm G}^{(0)}=\mathbf 0,\quad
 \bm{\theta}_\mathrm{C}^{(0)}=\begin{bmatrix} \bm{m^{(0)}} \\ \bm{k^{(0)}} \end{bmatrix}=\begin{bmatrix} \bm{1} \\ \bm{0} \end{bmatrix}, \quad
\bm{\theta}_{\mathrm T}^{(0)}=\mathbf 0,\quad
\bm{\theta}_{\mathrm J}^{(0)}=\mathbf 0.
\end{equation}
\section{Identification}
\label{sc:Identification}
\subsection{Problem Formulation}
\label{sc:ProblemFormulation}
The calibration is formulated as an optimization problem
$ 
    \label{eq:optimization}
    \min_{\bm{\theta}} L(\bm{\theta})
$ 
with objective $L(\bm{\theta})$ and parameter $\bm{\theta}$. The objective is the sum of squared position errors over all considered poses, measured as the Euclidean distance between the predicted and the measured tool position, and a regularization term:
\begin{equation}
    \label{eq:loss}
    L(\bm{\theta}) = \frac{1}{2\hat{m}} \sum_{m} \big\| \mathbf{t}^\mathrm{pred}(\mathbf{q}_m, \kappa_m; \bm{\theta}) - \mathbf{t}^\mathrm{meas}_m \big\|^2 + L_{\mathrm{reg}} (\bm{\theta}).
\end{equation}
In the resulting nonlinear least-squares problem, \( \mathbf{t}^\mathrm{pred} \) corresponds to the translational component of \( \bm{T}^{\mathrm{TCP}}_{\mathrm{glob}} \) from~\eqref{eq:FullRobotModel}. The subscript $m = 1, \dots, \hat{m}$ indicates the measured pose associated with the commanded joint states $\mathbf{q}_m$, the ambient temperature $\kappa_m$, and the measured tool position $\mathbf{t}^\mathrm{meas}_m$. We write $|| \cdot ||$ for the Euclidean norm. The regularization term is detailed in Sec.~\ref{sc:Regularization}.
\subsection{Optimization Algorithm}
\label{sc:OptimizationAlgorithm}
We solve the optimization problem using a regularized Gauss-Newton algorithm. This algorithm, along with its variations, is a commonly used method for solving nonlinear least-squares problems \cite{bjorck_numerical_1996} and a well-established choice for robot calibration \cite{mooring_fundamentals_1991}. The algorithm's update rule is given as:
\begin{equation}
    \label{eq:GaussNewtonUpdate}
    \bm{\theta}^{(k+1)} = \bm{\theta}^{(k)} - \left( B + \lambda_{\mathrm{GN}} \mathbb{I} \right)^{-1} \nabla_\theta L,
\end{equation}
for $k = 0, 1, \dots $, with initialization $\bm{\theta}^{(0)}$. We introduce $\lambda_{\mathrm{GN}}$ as a regularization parameter. The gradient, $\nabla_\theta L$, and the approximate Hessian, $B$, are detailed in the following.
For the calculation of these first and second derivatives we use the short-hand notation of $\mathbf{t}^\mathrm{pred}_m$ for $\mathbf{t}^\mathrm{pred} (\bm{q}_m, \kappa_m; \bm{\theta})$. 
Differentiation of~\eqref{eq:loss} yields the objective's gradient
\begin{equation}
    \label{eq:LossGradient}
    \nabla_\theta L = \frac{1}{\hat{m}} \sum_m \left( \mathbf{t}^\mathrm{pred}_m - \mathbf{t}^\mathrm{meas}_m \right)^\mathrm{T} \frac{\partial \mathbf{t}^\mathrm{pred}_m}{\partial \bm{\theta}} + \frac{\mathrm{d} L_{\mathrm{reg}}}{\mathrm{d} \bm{\theta}},
\end{equation}
and the objective's Hessian
\begin{equation}
    \label{eq:LossHessian}
\begin{split}
    \frac{\mathrm{d}^2 L}{\mathrm{d} \bm{\theta}^2} = & \frac{1}{\hat{m}} \sum_m \Big( \frac{\partial \mathbf{t}^\mathrm{pred}_m}{\partial \bm{\theta}} ^\mathrm{T} \frac{\partial \mathbf{t}^\mathrm{pred}_m}{\partial \bm{\theta}} + \\ & \left( \mathbf{t}^\mathrm{pred}_m - \mathbf{t}^\mathrm{meas}_m \right)^\mathrm{T} \frac{\partial^2 \mathbf{t}^\mathrm{pred}_m}{\partial \bm{\theta}^2} \Big)  + \frac{\mathrm{d}^2 L_{\mathrm{reg}}}{\mathrm{d} \bm{\theta}^2}.
\end{split}
\end{equation}
Based on this, we define the approximate Hessian used in the Gauss-Newton update by omitting the second-order term:
\begin{equation}
    B = \frac{1}{\hat{m}} \sum_m \frac{\partial \mathbf{t}^\mathrm{pred}_m}{\partial \bm{\theta}} ^\mathrm{T} \frac{\partial \mathbf{t}^\mathrm{pred}_m}{\partial \bm{\theta}} + \frac{\mathrm{d}^2 L_{\mathrm{reg}}}{\mathrm{d} \bm{\theta}^2}.
\end{equation}
This corresponds to approximating the forward kinematics linearly, which is reasonable since we work with small residuals. 
\subsection{Gradient Calculation}
\label{sc:GradientCalculation}
In the following, we summarize the calculation of the forward kinematics' gradient with respect to the model parameters. This can be done based on gradient propagation through the kinematic chain, given the individual gradient of each transformation model that are shown in Sec.~\ref{sc:CalibratedRobotModel}.
Except for link masses, all parameters affect only a single transformation in the kinematic chain. Link masses, on the other hand, affect all compliance transformations prior to that link in the kinematic chain. Therefore, we handle the case of link mass parameters separately.
We start with the case of the parameter $\theta_j$ not describing a link mass. Let $\bm{T}'$ be the transformation affected by $\theta_j$, and $\bm{T}'_{-}$ and $\bm{T}'_{+}$ be the transformations right before and right after $\bm{T}'$. Differentiating~\eqref{eq:FullRobotModel} yields
\begin{equation}
    \frac{\partial \boldsymbol{T}^\mathrm{TCP}_{\mathrm{glob}}}{\partial \theta_j} = \boldsymbol{T}^\mathrm{base} \dots \boldsymbol{T}'_{-} \frac{\partial \boldsymbol{T}'}{\partial \theta_j} \boldsymbol{T}'_{+} \dots \boldsymbol{T}^\mathrm{TCP}_\mathrm{local}.
    \label{eq:ChainDerivativeSimple}
\end{equation}
For mass parameters, this yields the summation 
\begin{equation}
    \frac{\partial \boldsymbol{T}^\mathrm{TCP}_{\mathrm{glob}}}{\partial \theta_j} = \sum_{\boldsymbol{T}' \in \{ \boldsymbol{T}^\mathrm{C}_1 \dots \boldsymbol{T}^\mathrm{C}_6 \} }
    \boldsymbol{T}^\mathrm{base} \dots \boldsymbol{T}'_{-} \frac{\partial \boldsymbol{T}'}{\partial \theta_j} \boldsymbol{T}'_{+} \dots \boldsymbol{T}^\mathrm{TCP}_\mathrm{local} .
    \label{eq:ChainDerivativeMass}
\end{equation}
\subsection{Regularization}
\label{sc:Regularization}
During the construction of our robot model, we have paid close attention to avoid overparameterizing the individual submodels for each error source to ensure their unique identifiability. However, unique identification is not guaranteed between the compliance model and the joint model, since both of them capture pose-dependent shifts in the joint position so that a certain overlap is likely. To eliminate the resulting nullspace, we apply a mild regularization to the joint model. In the construction of the regularization, we introduce a scaling, $l_i$, for each joint:
\begin{equation}
    L_\mathrm{reg}(\bm{\theta}) = \lambda_\mathrm{J} \sum_{i=1}^6 || \frac{1}{l_i} \bm{\theta}_{\mathrm{J}, i} ||^2.
\end{equation}
The scaling is performed according to the magnitude of joint errors we expect to see. Due to the cost-optimized manufacturing of industrial robots, every joint can be expected to be built to as much precision as necessary. Errors in joints near the base play into end-effector offsets more strongly than those in joints near the tool. Accordingly, we use the averaged sensitivity of the end-effector to each joint angle as a data-driven estimate of the necessary scaling, which we calculate using the unaugmented robot model~\eqref{eq:RobotModelUncalibrated}, as
\begin{equation}
    l_i = \frac{1}{\hat{m}} \sum_m \left| \left|  \frac{\partial \bm{t}^{\mathrm{TCP}}_{\mathrm{glob}}(\bm{q}_m)}{\partial q_i} \right| \right|.
\end{equation}
The regularization loss is then added to the loss over residuals as shown in~\eqref{eq:loss}. We set the hyperparameter $\lambda_\mathrm{J}$ to $=10^{-5}$, which proved to be a reliable choice throughout our experiments.
In addition to regularization of the loss, we also introduced a regularization of the optimization algorithm, as seen in~\eqref{eq:GaussNewtonUpdate}, which resembles a Levenberg-Marquardt damping \cite[chapter 5]{bard_nonlinear_1974}. The scaling factor $\lambda_{\mathrm{GN}}$ balances speed of convergence with robustness of the optimization. To tune this hyperparameter, we found it helpful to perform a coarse one-dimensional grid search. For this study, we use $\lambda_{\mathrm{GN}} = 10^{-7}$.
\section{Experimental Setup}
\label{sc:ExpSetup}
The experiments are performed with a KUKA KR30-3 industrial robot arm. This robot uses six revolute joints, has a maximum reach of 2033\si{\milli\metre}, and a rated payload of 30\si{\kilogram}. To emulate a more realistic application, a payload in the form of a dead weight was mounted on the robot's end-effector. The pose repeatability is given by the manufacturer as 60\si{\micro\metre}.
\begin{figure}[ht]
    \centering
    \includegraphics[width=\columnwidth, trim={0cm 0cm 0 0cm},clip]{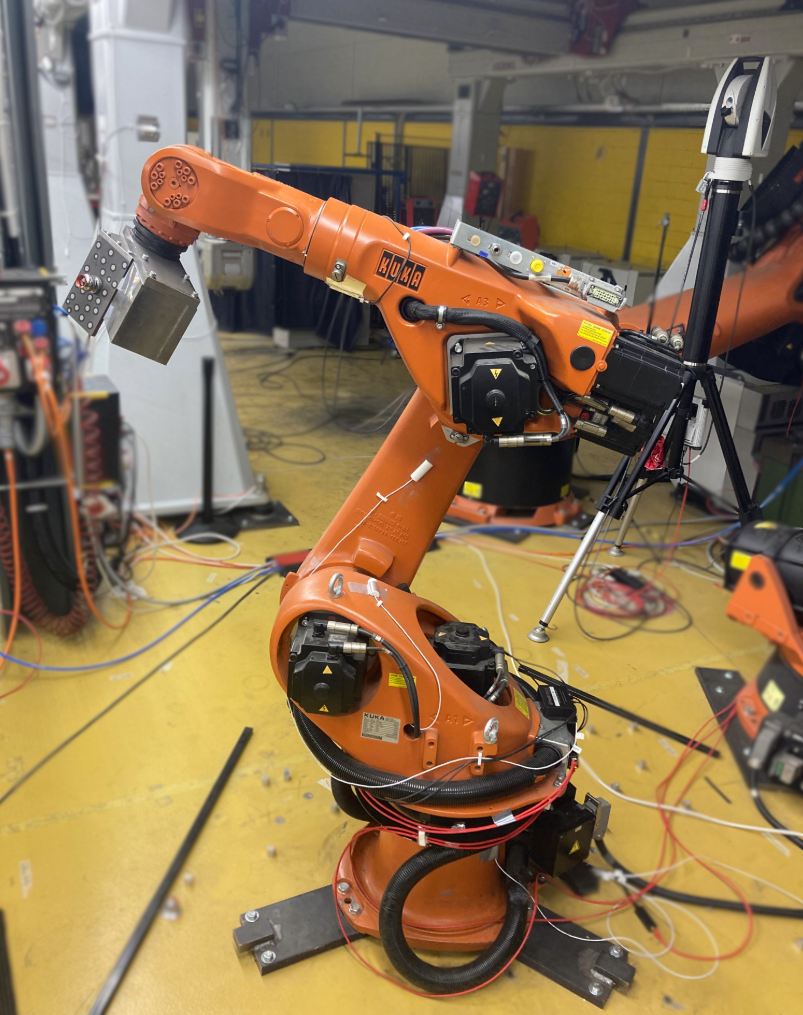}
    \caption{Lab setup: payload and reflector on the tool (left) and laser tracker (right).}
    \label{fig:LabSetup}
\end{figure}
The tool position is measured by a Leica AT960 laser tracker. The tracker's maximum permissible error per Cartesian axis is given as 15\si{\micro\metre} + 6\si[per-mode=symbol]{\micro\metre\per\metre}, with our measurements being taken at a distance of 2-3\si{\metre} between tracker and tool.
To account for the robot's specific load case, workspace, and potential wear, as well as the effective tracker noise, we will perform an experimental repeatability estimation as described in section~\ref{sc:Repeatability}.
\subsection{Data Collection}
\label{sc:DataCollection}
\begin{figure}[ht]
  \centering
  \includegraphics[width=\linewidth]{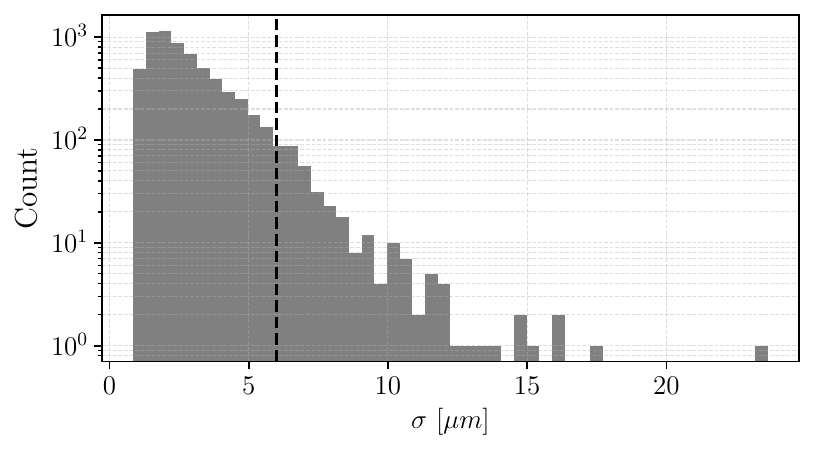}
  \caption{Histogram of the variation in laser tracker measurements. The dashed line indicates the applied threshold, effectively removing roughly 5.2\% of data points.}
  \label{fig:LaserTrackerReduction}
\end{figure}

\begin{figure}[ht]
  \centering
  \includegraphics[width=\linewidth]{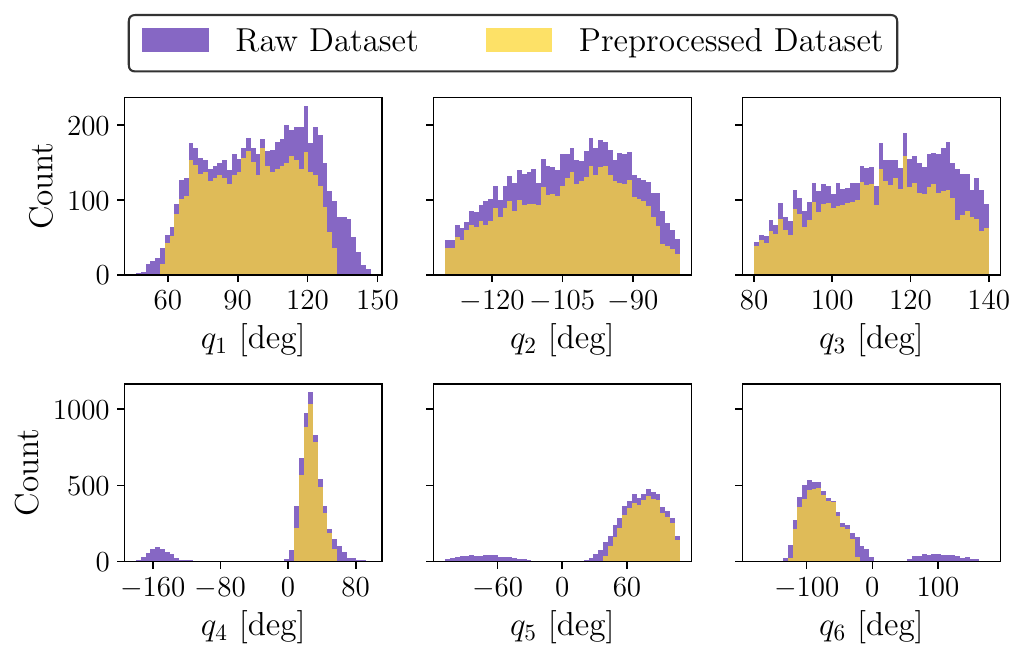}
  \caption{Dataset filtering to ensure a minimum amount of data per joint model support point.}
  \label{fig:DatasetReduction}
\end{figure}

\begin{figure}[ht]
    \centering
    \begin{subfigure}[b]{0.48\textwidth}
        \centering
        \begin{tikzpicture}
            \begin{axis}[
                    width=\textwidth, 
                    height=4cm,
                    xlabel={time [days]},
                    xtick = {0, 1, 2},
                    ylabel={$\kappa $ [°C]},
                ]
                \addplot[black] table [x expr=1/(3600*24)*\thisrow{timestamp}, y=T9, col sep=comma] {data_full_Temp_only.tex};
            \end{axis}
        \end{tikzpicture}
    \end{subfigure}%
    \caption{Ambient temperature during the experiment. Note that the discrete levels are due to the inherent resolution of the temperature sensor used.}
    \label{fig:TemperatureEvolutionData}
\end{figure}
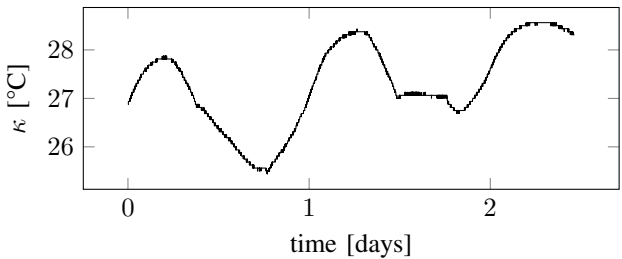

Measurement poses are sampled uniformly over a cartesian workspace of dimension $1200\si{\milli \metre} \times 700 \si{\milli \metre} \times 900 \si{\milli \metre}$. The orientation of the tool is set to face the tracker, and then artificially offset by a random rotation around each principal axis. Each offset is sampled from a uniform distribution in $[-15\si{\degree}, 15\si{\degree}]$. We filter out locations where the inverse kinematic leads to solutions outside the robot's joint constraints or where the tool is invisible to the laser tracker. All poses are approached from the same direction in joint space to eliminate errors caused by backlash. The environmental temperature during the experiment, measured near the robot, is shown in~Fig.\ref{fig:TemperatureEvolutionData}. Before recording poses, we warm up the robot by performing movements similar to those used during the experiment. Thermal steady state is confirmed using a thermal camera.
The precision of laser tracker measurements is subject to various factors. Apart from characteristics of the experimental setup and procedure, such as the distance between the tracker and the reflector, this further includes the mechanical stability of the entire setup, or the inclusion of a warm-up phase for the tracker. Several aspects of the environmental conditions have also been demonstrated to affect the measurement precision negatively via their effect on the air's refractive index \cite{ciddor_refractive_1996, tyler_estler_high-accuracy_1985}: temperature, pressure, CO$_2$ concentration, humidity, or air turbulence \cite{estler_large-scale_2002, perez_munoz_analysis_2016}. All of these effects are difficult to predict or compensate for and would necessitate additional sensor information or a controlled experimental environment. We therefore discard any measurement whose within-sample variability, $\sigma$, exceeded a predetermined threshold of $6\si{\micro \metre}$. 
The distribution of $\sigma$ over the raw measurements is visualized in Fig.~\ref{fig:LaserTrackerReduction}.
The joint model we propose requires data points distributed across all relevant segments of each joint’s active range to ensure proper identification. To fulfill this requirement, we deem the joint model, and thus the entire robot model, to only be applicable in regions where enough data is available. Specifically, we require at least 10 data points for each piece-wise section of the joint model, as introduced in~\eqref{eq:JointCorrection}, to properly identify all model parameters for all joints. Based on this requirement and the available data we conclude that the resolution of our joint model~\eqref{eq:JointCorrection} that can be achieved over all joints is $d_\mathrm{supp}=80$ \si{\per\rad}. As a consequence of these constraints, any data outside the range where the minimal density of data points is available becomes unusable and is discarded. The corresponding dataset reduction is visualized in Fig.~\ref{fig:DatasetReduction}. In total, a number of $\hat{m} = 4571$ poses remain in our dataset, which we denote as
\begin{equation} \mathcal{D} = \{ (\bm{q}_1, \kappa_1, \bm{t}^{\mathrm{meas}}_1), \dots, (\bm{q}_{\hat{m}}, \kappa_{\hat{m}}, \bm{t}^{\mathrm{meas}}_{\hat{m}}) \}.
\end{equation} 

\section{Model Validation}
\label{sc:ModelValidation}
We have introduced three main requirements to our calibration method: accuracy, robustness, and practical applicability. 
While the practical applicability is a byproduct of our model design, as it assumes no prior knowledge and a simple experimental design, the accuracy and robustness remain to be demonstrated. In the following, we present the experiments we perform to confirm these properties.
\paragraph{Accuracy}
To assess the accuracy that our method achieves, we perform a number of calibrations according to a fold-wise cross-validation principle. Specifically, we split the entire dataset into five consecutive chunks in time and perform five individual calibrations, each using four out of the five chunks as the training set. The remaining chunk constitutes the validation set. The cross-validation structure allows for robust performance estimation, whereby the temporal split minimizes the leakage of potential time-dependent effects. 
We carry out the individual identification procedures as described in Sec.~\ref{sc:Identification} and initialized with $\bm{\theta}^{(0)}$ as defined in~\eqref{eq:InitialGuess}. Our primary metric for the error associated with an individual pose with index $m$ is the Euclidean distance between the measured and the predicted end-effector position, i.e., the norm of the residual:
\begin{equation}
    \label{eq:PoseError}
    e_m = \big\| \mathbf{t}^\mathrm{meas}_m  - \mathbf{t}^\mathrm{pred}_m \big\|.
\end{equation}
As a measure of the mean error, we will use the average over the training residuals of all folds and the average over the validation residuals of all folds. As a measure of the worst-case error for each fold, we calculate the 95\% percentile and maximum error and then report the maximum for each over all folds. Again, this is done once for the train sets and once for the validation sets. As a result, we have a reliable metric for both the mean error we can expect from our model and its worst-case performance.
We place these accuracy estimates in relation with the inherent variability of our setup, namely the combined effects of the robot and laser tracker, which provides a lower bound for the accuracy we may hope to achieve. Sec.~\ref{sc:Repeatability} presents this procedure.
\paragraph{Robustness}
Firstly, we consider robustness to measurement noise. One aspect of robustness to noise is the numerical conditioning of the optimization, which we investigate in Sec.~\ref{sc:SVDAnalysis}. 
Secondly, we investigate the performance using reduced training data, presented in Sec.~\ref{sc:TraingDataSensitivity}. This experiment helps assess whether we have enough data points to fully identify the model we suggest with the data we have. Additionally, this test may be of practical relevance whenever downtime for calibration may be costly, and the trade-off between experiment duration and calibration accuracy thus needs to be balanced.
Finally, we demonstrate the robust interaction between the joint model and the remaining part of the model. This submodel is of special interest as it has by far the largest number of free parameters (one might consider its capabilities similar to those of a residual model as described in Sec.~\ref{sc:Regularization}), which could raise concerns of interference with other submodels. In Sec.~\ref{sc:JointCorrectionIdentification} we address these concerns by comparing the residuals of a calibration without joint model to the learned joint model.
\begin{table*}[ht]
\sisetup{detect-weight}
\centering
\renewcommand{\arraystretch}{0.9}
\begin{tabularx}{\textwidth}{%
  >{\raggedright\arraybackslash}X
  *{3 }{S[round-mode=places,round-precision=1]}
  *{3}{S[round-mode=places,round-precision=1]}
}
\toprule
\textbf{Model} & \multicolumn{3}{c}{\textbf{Training error [$\mu$m]}} & 
                 \multicolumn{3}{c}{\textbf{Validation error [$\mu$m]}} \\
\cmidrule(r){2-4} \cmidrule(l){5-7}
 & \textbf{Mean} & \textbf{95\%} & \textbf{Max} & 
   \textbf{Mean} & \textbf{95\%} & \textbf{Max} \\
\midrule
geometry & 99.324312 & 194.699599 & 393.733854 & 102.339096 & 199.231545 & 384.342813 \\
geometry + compliance & 65.654994 & 117.371864 & 197.332842 & 69.941065 & 125.020599 & 198.354063 \\
geometry + compliance + thermal & 57.845036 & 104.471756 & 188.279358 & 58.463990 & 105.613716 & 189.057085 \\
full model & 24.499717 & 47.315902 & 97.623119 & \B 26.768233 & \B 50.955568 & \B 97.422424 \\
\bottomrule
\end{tabularx}
\caption{Five-fold temporal cross-validation: mean, 95th-percentile, and maximum position errors for ablative model variants.}
\label{fig:ModelComparison}
\end{table*}
\section{Repeatability Estimation}
\label{sc:Repeatability}
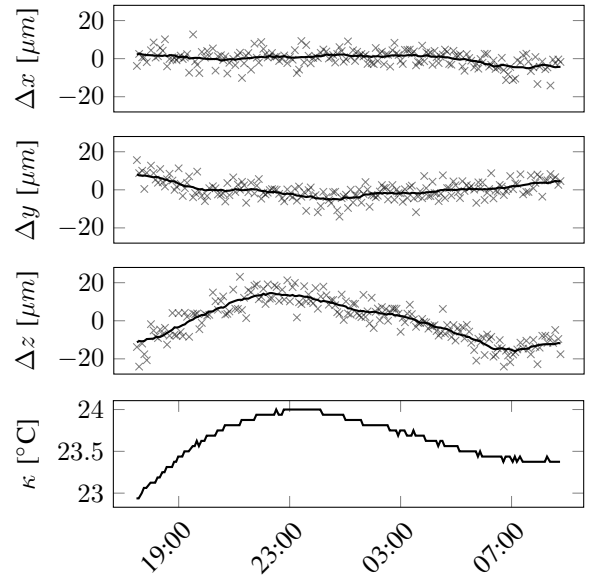
\begin{figure}[ht]
\centering
\pgfplotsset{
    every axis/.append style={
        width=0.9\textwidth, height=3cm,
        xlabel={}, ylabel near ticks,
        xticklabel style={rotate=45, anchor=south, yshift=-0.5cm, xshift=-0.5cm},
    }
}
    \begin{subfigure}[b]{0.48\textwidth}
      \centering
        \begin{tikzpicture}
        \centering
        \begin{axis}[
            ylabel={$\Delta x$ [$\mu m$]},
            xlabel=\empty,
            xtick=\empty,
            ymax=28, ymin=-28,
            xmin=-3000, xmax=5480 + 14.6*3600,
            ]
            \addplot[only marks, mark=x, opacity=0.5, color=black] table[x=time_in_seconds, y=err_x, col sep=comma, header=true] {data_p1_analysis.tex};
            \addplot[thick, color=black] table[x=time_in_seconds, y=rolling_err_mean_x, col sep=comma, header=true] {data_p1_analysis.tex};
        \end{axis}
        \end{tikzpicture}
    \end{subfigure}
    \begin{subfigure}[b]{0.48\textwidth}
        \centering
        \begin{tikzpicture}
        \centering
        \begin{axis}[
            ylabel={$\Delta y$ [$\mu m$]},
            xlabel=\empty,
            xtick=\empty,
            ymax=28, ymin=-28,
            xmin=-3000, xmax=5480 + 14.6*3600,
            ]
            \addplot[only marks, mark=x, opacity=0.5, color=black] table[x=time_in_seconds, y=err_y, col sep=comma, header=true] {data_p1_analysis.tex};
            \addplot[thick, color=black] table[x=time_in_seconds, y=rolling_err_mean_y, col sep=comma, header=true] {data_p1_analysis.tex};
        \end{axis}
        \end{tikzpicture}
    \end{subfigure}
    \begin{subfigure}[b]{0.48\textwidth}
        \centering
        \begin{tikzpicture}
        \centering
        \begin{axis}[
            ylabel={$\Delta z$ [$\mu m$]},
            xlabel=\empty,
            xtick=\empty,
            ymax=28, ymin=-28,
            xmin=-3000, xmax=5480 + 14.6*3600,
            ]
            \addplot[only marks, mark=x, opacity=0.5, color=black] table[x=time_in_seconds, y=err_z, col sep=comma, header=true] {data_p1_analysis.tex};
            \addplot[thick, color=black] table[x=time_in_seconds, y=rolling_err_mean_z, col sep=comma, header=true] {data_p1_analysis.tex};
        \end{axis}
        \end{tikzpicture}
    \end{subfigure}
    \begin{subfigure}[b]{0.48\textwidth}
        \centering
        \begin{tikzpicture}
        \centering
        \begin{axis}[
            at={(0,-12cm)}, anchor=north west,
            xlabel=\empty,
            ylabel={$\kappa$ [$^\circ$C]},
            xtick={5.480, 5.480 + 4*3.600, 5.480 + 8*3.600, 5.480 + 12*3.600},
            xticklabels={19:00, 23:00, 03:00, 07:00},
            xmin=-3.000, xmax=5.480 + 14.6*3.600,
        ]
    \addplot[thick, color=black] table[x expr=0.001*\thisrow{time_in_seconds}, y=T1, col sep=comma, header=true] {data_p1_analysis.tex};
    \end{axis}
    \end{tikzpicture}
\end{subfigure}
\caption{Repeated measurements of the end-effector position for the same joint state. Shown is the deviation of each coordinate from its mean, along with the ambient temperature over time.}
\label{fig:repeatability}
\end{figure}
The hardware specifications of the robot and the tracker indicate a total repeatability of above $60\si{\micro\metre}$. Since we will find calibration results with mean distance errors well below this value, we re-estimate the setup's effective repeatability. It needs to be taken into account that the manufacturer-provided repeatability values should generally be considered a confidence range under specific challenging conditions rather than a measure for the effective spread, especially under stable experimental conditions. Therefore, we will not consider the setup's nominal repeatability as the benchmark of the calibration, but the experimentally determined repeatability presented in the following.
We follow the procedure of ISO 9283:1998 for the assessment of the robot's repeatability. This consists of randomly sampling a list of joint states and repeatedly iterating through this list while measuring the tool position at each stop, producing a cluster of position measurements for each considered pose. The position measurements are referred to as $\bm{t}_{i, j}$, for each point $j = 1, \dots, n_\mathrm{pt}$ in each cluster $i = 1, \dots, n_{\mathrm{cl}}$. The cluster centers and each point's distance to its cluster center, respectively, are 
\begin{equation}
    \bar{\bm{t}}_i = \frac{1}{n_\mathrm{pt}} \sum_{j} \bm{t}_{i,j}
    \quad \text{and} \quad
    l_{i,j} = \left\| \bm{t}_{i,j} - \bar{\bm{t}}_i \right\|.
\end{equation}
The average distance of points to their cluster center and the standard deviation of this distance are
\begin{equation}
    \label{eq:average_error}
    \bar{l} = \frac{1}{n_\mathrm{pt} \cdot n_\mathrm{cl}} \sum_{i, j} l_{i,j} \text{ and } S_l = \sqrt{ \frac{\sum_{i, j} (l_{i,j} - \bar{l})^2 }{n_\mathrm{pt} \cdot n_\mathrm{cl} - 1} }.
\end{equation}
This yields the repeatability
\begin{equation}
    \label{eq:repeatability}
    RP = \bar{l} + 3 S_{l}.
\end{equation}
We run a warm-up procedure for roughly five hours until we can confirm that the thermal effects caused by the robot's movements have reached a steady state. Following this, measurements were taken over one night for roughly 14 hours. Fig.~\ref{fig:repeatability} shows the offset of position measurements for one of the selected poses to their overall average for each direction: $\Delta x$, $\Delta y$, and $\Delta z$. We identify two main effects: a slow drift over time and quasi-random variations. Due to the visible correlation of the drift to the ambient temperature measurement (similar trends hold for all analyzed poses), we attribute the drift to thermal effects. 
Ideally, we would like to characterize the system's repeatability without external effects such as changes in the ambient temperature, but since we cannot guarantee a controlled environment in our setup, we estimate the repeatability by compensating for the observed drift. We do this by applying a rolling mean to the data (the solid line in Fig.~\ref{fig:repeatability}) and subtracting it from the original data. Further we discard a small number of measurements in the beginning and end where the drift estimate is poor. From the corrected measurements, we extract the following estimates for the repeatability using~\eqref{eq:average_error} and~\eqref{eq:repeatability}: $\bar{l} = 6.26 \si{\micro \metre}$, $S_l = 2.75 \si{\micro \metre}$, and $RP = 14.52 \si{\micro \metre}$. 
\section{Results}
\label{sc:Results}
\subsection{Absolute Accuracy}
\label{sc:AbsoluteAccuracy}
Tab.~\ref{fig:ModelComparison} presents the main result of this study: the accuracy that our method achieves, highlighted in bold for our full suggested model. The table further provides a model ablation that compares fully converged calibration results when sequentially including additional submodels. All results presented here stem from a cross-validation experiment with systematic folds in time, as described in Sec.~\ref{sc:ModelValidation}, to produce reliable metrics for mean and worst-case errors. Tab.~\ref{tab:IdentifiedParameterValues} lists the identified parameter values for the \textbf{full model}, together with the deviation between folds.
\subsection{Identified Joint Model}
\label{sc:JointCorrectionIdentification}
\begin{figure*}[ht]
  \centering
  \includegraphics[width=\linewidth]{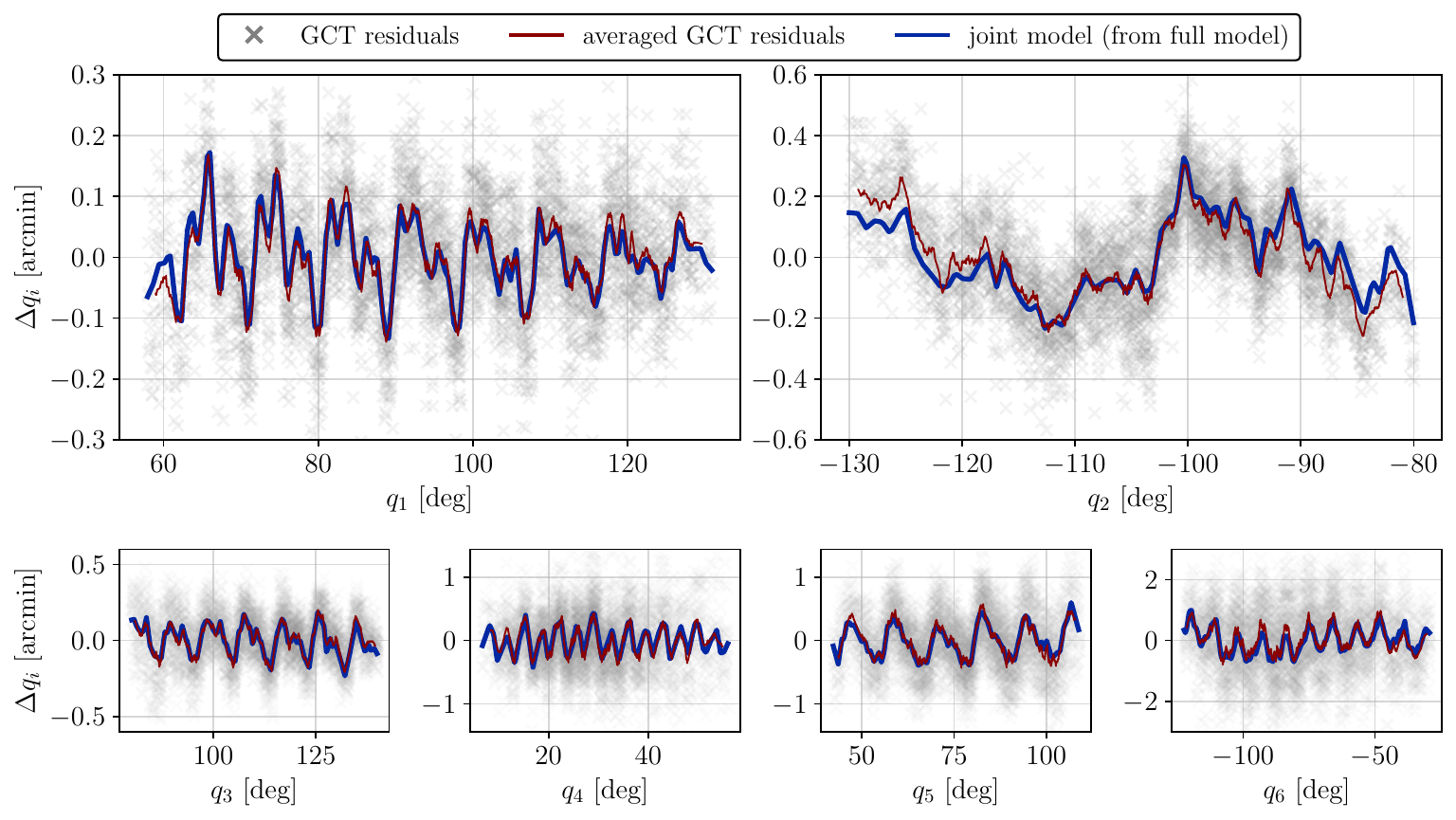}
    \caption{Joint model identified by the \textbf{full model} compared to residuals of the \textbf{GCT model} (no joint correction). The residuals are presented both raw and with a moving average filter.}
  \label{fig:JointIdentification}
\end{figure*}

As introduced in Sec.~\ref{sc:VirtualJointModeling}, our joint model is a piece-wise linear function relating commanded joint angles to real joint angles. Fig.~\ref{fig:JointIdentification} visualizes the resulting model of the joint behavior as identified by the \textbf{full model}. We compare the identified joint model to the measured joint errors. These measurements are derived from the remaining residual position errors after a calibration using the \textbf{GCT model} on the same data by projecting the residuals into joint space. We achieve this using the model's Jacobian so that for each residual $\mathbf{t}^\mathrm{meas}_m - \mathbf{t}^\mathrm{pred}_m$ and joint $i$ we calculate
\begin{equation}
    \bm{J}_{m,i} = \frac{\partial \mathbf{t}^\mathrm{pred}_m}{\partial q_i} \quad \text{and} \quad \Delta q_{m,i} = \frac{\bm{J}_{m,i}^\mathrm{T} \left( \mathbf{t}^\mathrm{meas}_m - \mathbf{t}^\mathrm{pred}_m \right) }{\bm{J}_{m,i}^\mathrm{T} \bm{J}_{m,i}}.
\end{equation}
\subsection{Sensitivity to Number of Training Poses}
\label{sc:TraingDataSensitivity}
\begin{figure}[ht]
    \centering
    \includegraphics[width=\linewidth]{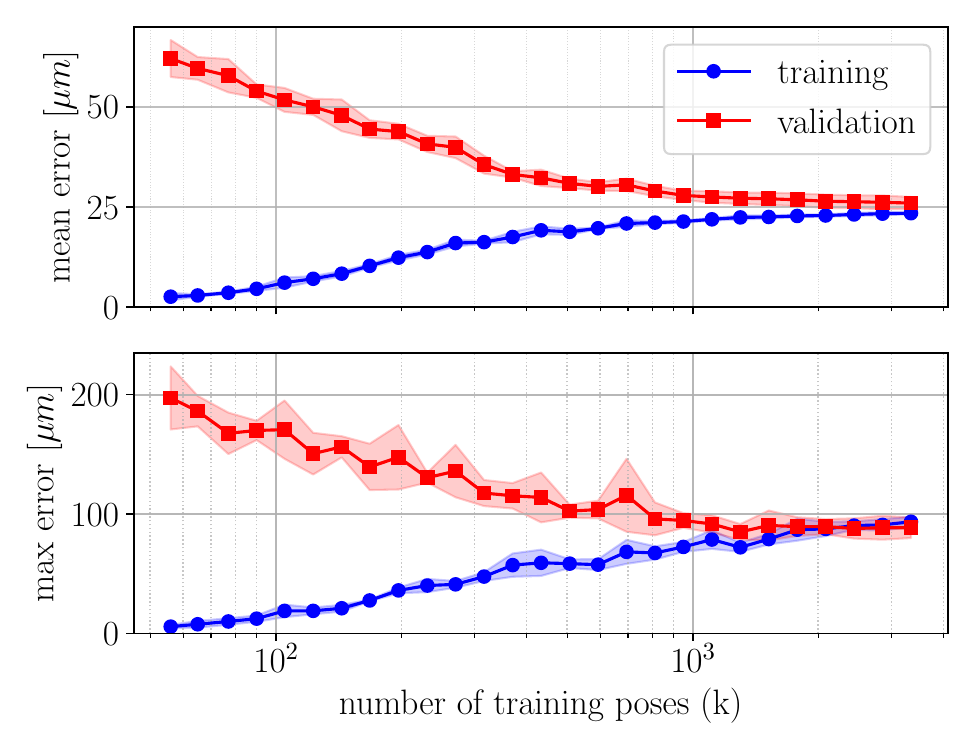}
    \caption{Calibration accuracy using reduced training data and the \textbf{full model}. The shaded area indicates the standard deviation over the temporal cross-validation.}
    \label{fig:TrainDataReduction}
\end{figure}

The next experiment investigates the model performance when reducing the number of training poses. We do this by repeating the previously used systematic fold-wise splitting and then randomly subsampling the training sets to a specific size, $k$, which we let be evenly spaced on a logarithmic scale. The validation sets remain unchanged. The resulting accuracies are presented in Fig.~\ref{fig:TrainDataReduction}.
\subsection{Local Submodel Gradient Structure}
\label{sc:SVDAnalysis}
\begin{figure}[t]
\centering
\pgfplotstableread[col sep=comma]{data_singular_values_summary.tex}\svdtable
\begin{tikzpicture}
\begin{groupplot}[
    group style = {group size = 1 by 4, vertical sep = 1.4em},
    width  = 0.98\linewidth,
    height = 0.40\linewidth,
    ymode  = log,
    every axis plot/.append style = {only marks, mark size = 3pt},
    grid = both,
    grid style = {dash pattern = on 1pt off 2pt},
    ylabel style={rotate=-90},
    xmin=0]
\nextgroupplot[ylabel = {$\sigma_i^{\mathrm{G}}$}]
  \addplot+[mark=x] table[x expr=\thisrow{component_index}+1, y = Kinematic] {\svdtable};
\nextgroupplot[ylabel = {$\sigma_i^{\mathrm{C}}$}]
  \addplot+[mark=x] table[x expr=\thisrow{component_index}+1, y = Compliance] {\svdtable};
\nextgroupplot[ylabel = {$\sigma_i^{\mathrm{T}}$}]
  \addplot+[mark=x] table[x expr=\thisrow{component_index}+1, y = Thermal] {\svdtable};
\nextgroupplot[ylabel = {$\sigma_i^{\mathrm{J}}$},
               xlabel = {Component index ($i$})]
  \addplot+[mark=x] table[x expr=\thisrow{component_index}+1, y = Joint] {\svdtable};
\end{groupplot}
\end{tikzpicture}
\caption{Singular-value spectra of the four submodel gradients near convergence.}
\label{fig:svd-spectra-stacked}
\end{figure}
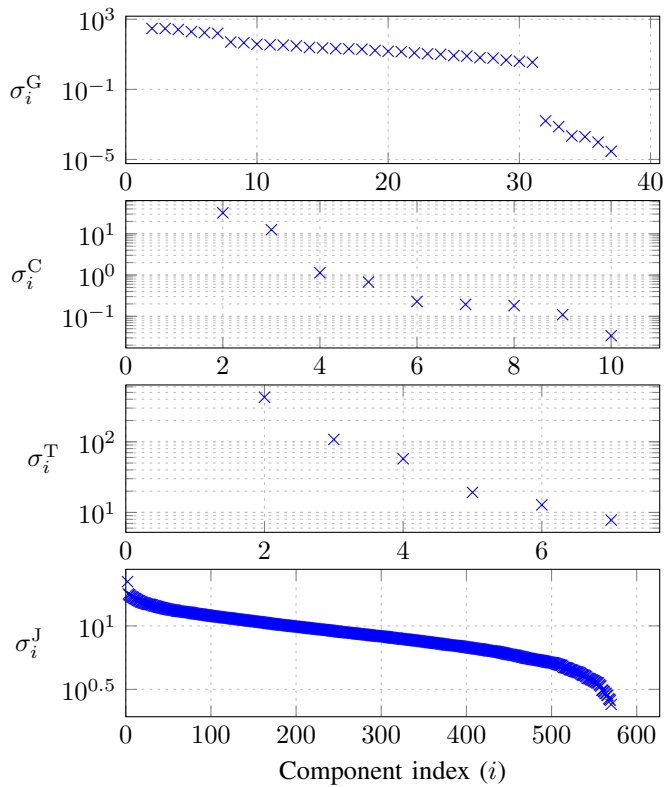
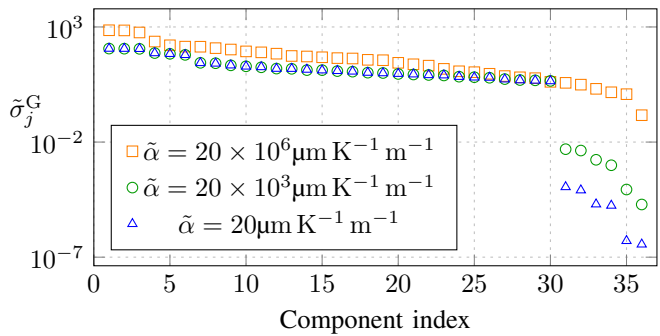
\begin{figure}[ht]
\centering
\begin{tikzpicture}
\begin{axis}[
    xlabel={Component index},
    ylabel={$\tilde{\sigma}_j^\mathrm{G}$},
    ylabel style={rotate=-90},
    y label style={at={(axis description cs:-0.12,.5)},anchor=south},
    ymode=log,
    grid=both,
    grid style = {dash pattern = on 1pt off 2pt},
    width=0.5\textwidth,
    height=5cm,
    mark options={scale=1.2},
    legend style={at={(0.03,0.06)}, anchor=south west, inner sep=1pt, row sep=0.1pt,
    inner xsep=6pt,  
    inner ysep=3pt,  
    },
    legend columns=1,
    xmin=0,
    xmax=37
]
\pgfplotstableread[col sep=comma]{data_SyntheticData_singular_value_spectra_T_10_to_50.tex}\svdata
\addplot+[
    only marks,
    mark=square,
    mark options={fill=orange, draw=orange},
    color=orange,
] table[x=SV_index, y=TC_2e+01] \svdata;
\addlegendentry{$\tilde{\alpha} = 20 \times 10^6\si{\micro \metre \per \kelvin \per \metre}$}
\addplot+[
    only marks,
    mark=o,
    mark options={fill=green!60!black, draw=green!60!black},
    color=green!60!black,
] table[x=SV_index, y=TC_2e-02] \svdata;
\addlegendentry{$\tilde{\alpha} = 20 \times 10^3\si{\micro \metre \per \kelvin \per \metre}$}
\addplot+[
    only marks,
    mark=triangle,
    mark options={fill=blue, draw=blue},
    color=blue,
] table[x=SV_index, y=TC_2e-05] \svdata;
\addlegendentry{$\tilde{\alpha} = 20 \si{\micro \metre \per \kelvin \per \metre}$}
\end{axis}
\end{tikzpicture}
\caption{Singular value spectra of geometry submodel gradient near convergence based on synthetic data for different thermal expansion coefficients $\tilde{\alpha}$.}
\label{fig:synthetic_data_SV_spectra}
\end{figure}
To assess the numerical stability and parameter identifiability of our model, we analyze the structure of the optimization gradients. In particular, we investigate whether any submodels suffer from local rank deficiencies that could impair reliable calibration.
We do this by constructing a per-sample gradient matrix for each submodel whose columns are the loss-gradient vectors, see~\eqref{eq:LossGradient}, so that
\begin{equation}
    \bm{g}_{m, \tilde{\bm{\theta}}} =  \left( \mathbf{t}^\mathrm{pred}_m - \mathbf{t}^\mathrm{meas}_m \right)^T \frac{\partial \mathbf{t}^\mathrm{pred}_m}{\partial \tilde{\bm{\theta}}}, \; \text{for} \;  \tilde{\bm{\theta}} \in \{ \bm{\theta}_{\mathrm{G}}, \bm{\theta}_{\mathrm{C}}, \bm{\theta}_{\mathrm{T}}, \bm{\theta}_{\mathrm{J}} \},
\end{equation}
can be collected into
\begin{equation}
    G_{\tilde{\bm{\theta}}} = \left[ \; g_{1, \tilde{\bm{\theta}}}^T \; g_{2, \tilde{\bm{\theta}}}^T \dots \; g_{\hat{m}, \tilde{\bm{\theta}}}^T \; \right].
\end{equation}
For the gradient matrix of each submodel, we perform a singular value decomposition, producing the four spectra of singular values $\bm{\sigma}^\mathrm{G}$, $\bm{\sigma}^\mathrm{C}$, $\bm{\sigma}^\mathrm{T}$, $\bm{\sigma}^\mathrm{J}$. This procedure can be tied to a geometrical argument for the retrieval of collinearities within the gradient \cite{g_w_stewart_collinearity_1987} or to a topological argument \cite[chapter 10.2]{nocedal1999numerical} considering that we approximate the Hessian using $G^TG$ (the empirical Fisher information). We show the sorted singular values in Fig.~\ref{fig:svd-spectra-stacked}.
As can be seen, most spectra show uniformly large singular values, indicating a well-conditioned loss surface and no rank loss of the corresponding gradients. However, in the spectrum of $G_\mathrm{G}$, we observe a sharp drop for the last six singular values, resembling a loss of rank. We closely investigate this pattern in Sec.~\ref{sc:SyntheticDataSVD} to rule out this possibility.
\section{Discussion}
\label{sc:Discussion}
\subsection{Thermal Influence on Geometric Identifiability}
\label{sc:SyntheticDataSVD}
As we have seen, the spectrum of singular-values corresponding to the geometric submodel had a potentially critical drop for the last six values. 
We hypothesize that the influence of the thermal effects on the identifiability of the geometric submodel can explain this. As introduced, our geometric model considers five parameters per joint, compared to four parameters in a standard DH-type parametrization to account for thermally varying link transformations. It therefore seems likely that the importance of this additional parameter changes with the magnitude of the thermal expansion. In Fig.~\ref{fig:synthetic_data_SV_spectra}, we support this hypothesis with a brief numerical experiment with synthetic data, where we artificially scale the thermal expansion coefficient. Specifically, we investigate the geometric submodel's singular-values of three calibrations using synthetic datasets. For every dataset we generate $1000$ random data points, consisting of poses and ambient temperatures with end-effector positions predicted by the calibrated \textbf{full model} (augmented with the assumed thermal expansion coefficient in all thermal transformations). The thermal expansion coefficients were artificially set to either $\tilde{\alpha} = 20 \si{\micro \metre \per \kelvin \per \metre}$, a physically plausible value, or scaled versions thereof as seen in Fig.~\ref{fig:synthetic_data_SV_spectra}.
This experiment shows how the critical drop off is initially reproduced but shrinks and ultimately vanishes with increased magnitude of the thermal effects.
\subsection{Accuracy and Robustness}
\label{sc:Disc_AccuracyRobustness}
\begin{figure}[ht]
    \centering
    \includegraphics[width=\linewidth]{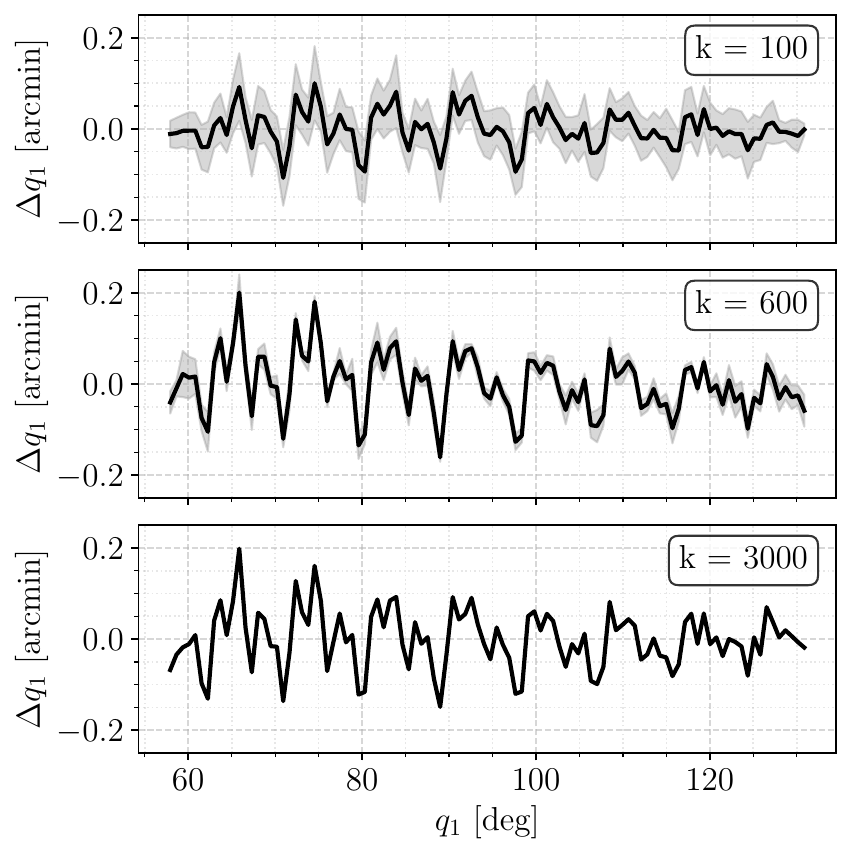}
    \caption{Joint 1 correction curves for $k=100, 600, 3000$ training poses. For cases with several folds ($k=100, k=600$), we present the mean over folds and the standard deviation between folds.}
    \label{fig:JointModelForReducedTrainSets}
\end{figure}

The goal of our calibration is the derivation of an accurate model of the forward kinematics of a specific articulated robot. Tab.~\ref{fig:ModelComparison} demonstrates that we have achieved this goal by reducing remaining errors to an average of $26.8 \si{\micro \metre}$ and a maximum of $97.4 \si{\micro \metre}$. Compared to the average error of a quasi-uncalibrated robot (determining only the robot base and the tool dimension) of roughly 7.6 $\si{\milli \metre}$ for the same data, this is an improvement of more than 99\%.
As shown in Tab.~\ref{fig:ModelComparison}, errors decrease as additional submodels are included. This shows that each submodel effectively captures additional error sources. Further, we see good generalization properties across the board, indicated by a validation error that is only slightly higher than the training error. This indicates that we do not significantly overfit either on measurement noise, or (since we have validated on temporally separate data) on temporal dynamics such as ambient thermal effects.
While the physicality of the model parameters has not been a target of our calibration (we have leveraged physical assumptions exclusively for their benefits of data efficiency and reduced modes for overfit), we still suggest analyzing the retrieved parameters to investigate their variation over folds. Tab.~\ref{tab:IdentifiedParameterValues} presents the parameters of the compliance submodel and the thermal submodel as found by the \textbf{full model} with an indication for their standard deviation over the folds used by our cross-validation scheme. Observing the thus computed values, we conclude that the parameters were identified robustly. The relatively small variation of identified parameter values, listed in Tab.~\ref{tab:IdentifiedParameterValues}, demonstrates the robustness of the model identification from data. Note that the retrieved parameter values of the compliance model inherently depend on the choice for the center of mass locations, $r_1, \dots r_6$. In some cases, this even results in negative mass values. To achieve physicality of the identified mass values, accurate center of mass information would be a requirement.
Considering the experiment with less training data, shown in the figure, we briefly investigate the origin of the larger generalization error visible in Fig.~\ref{fig:TrainDataReduction} for the reduced training data. Since the joint model has the largest number of free parameters, it is subject to the highest risk of overfitting. We consider a training set of 3000 poses, randomly drawn from our entire dataset, and investigate the learned joint model for joint 1 under the following three settings: a single calibration using all $k = 3000$ poses; five calibrations using random partitions of the training set into five disjoint subsets of $k = 600$ poses each, and 30 calibrations using random partitions into 30 disjoint subsets of $k = 100$ poses each.
In Fig.~\ref{fig:JointModelForReducedTrainSets}, we compare the resulting models. As expected, we observe a larger variation in the models that were trained on less data. We also see a tendency for more conservative joint models, i.e. corrections generally closer to zero, when less data is available (most visible in the $k=100$ case), which we attribute to the regularization.
\begin{table}[ht]
  \centering
  \renewcommand{\arraystretch}{0.9}
  \begin{tabular}{>{\hspace{6em}}l r @{\hspace{0.15em}} c @{\hspace{0.15em}} l}
  \toprule
  \multicolumn{4}{c}{\textbf{lumped mass parameters} [\(\times\) \si{\kilogram}]} \\
  \midrule
  $m_2$ &  67.0 & \(\pm\) & 0.76 \\
  $m_3$ & -13.2 & \(\pm\) & 1.30 \\
  $m_4$ &  34.6 & \(\pm\) & 0.55 \\
  $m_5$ & -26.5 & \(\pm\) & 0.25 \\
  $m_6$ &   1.0 & \(\pm\) & 0 (fixed) \\
  \addlinespace
  \toprule
  \multicolumn{4}{c}{\textbf{joint compliance parameters} [\(\times 10^{-6}\) \si{\radian\per{\newton\metre}}]} \\
  \midrule
  $k_2$ &  3.87 & \(\pm\) & 0.40 \\
  $k_3$ &  4.26 & \(\pm\) & 0.45 \\
  $k_4$ & 20.3  & \(\pm\) & 0.61 \\
  $k_5$ & 47.8  & \(\pm\) & 0.23 \\
  $k_6$ & 70.1  & \(\pm\) & 0.52 \\
  \addlinespace
  \toprule
  \multicolumn{4}{c}{\textbf{thermal expansion parameters} [\(\times\) \si{\micro\metre\per\kelvin\per\metre}]} \\
  \midrule
  $\alpha_{1}$ & 28.85 & \(\pm\) & 0.41 \\
  $\alpha_{2}$ & 32.43 & \(\pm\) & 0.47 \\
  $\alpha_{3}$ & 13.45 & \(\pm\) & 3.54 \\
  $\alpha_{4}$ & 18.98 & \(\pm\) & 0.25 \\
  $\alpha_{5}$ & 21.36 & \(\pm\) & 2.71 \\
  $\alpha_{6}$ & 30.33 & \(\pm\) & 1.29 \\
  \bottomrule
  \end{tabular}
  \caption{The identified parameter values (mean \(\pm\) standard deviation
           over cross‐validation folds). Note that negative mass parameters are a consequence of the canonical choice for the center of mass locations.}
  \label{tab:IdentifiedParameterValues}
\end{table}
As we see in Fig.~\ref{fig:JointIdentification}, the \textbf{full model}'s joint behavior and the averaged residuals of the \textbf{GCT model} greatly coincide. This suggests that the piece-wise linear function can model the joint behavior sufficiently well and that its parameters have been properly identified from the data. 
An alternative interpretation of the joint model would be to draw comparisons to residual models.
Even though our joint model is identified in a combined optimization with all other submodels, Fig.~\ref{fig:JointIdentification} demonstrates that it captures only errors that the rest of the model cannot account for (those errors being what residual models would aim to capture). This is a strong indication that the staged learning setting of traditional residual models might not always be necessary for accurate calibration. We attribute this to the informed design of our joint model compared to an uninformed application of a general-purpose residual model. Whereas many of the recently studied residual models map from the joint space onto errors in cartesian space using black-box techniques \cite{chen_evolutionary_2020, zhao_system_2019, nguyen_calibration_2015}, which requires the model to learn complex representations and dependencies, including the robot kinematics themselves, our joint model has to learn only 6 individual $\mathbb{R} \rightarrow \mathbb{R}$ mappings. This promises a significant increase in data efficiency.
Note further that the identified mappings between commanded and actual angles are in line with previous research in terms of overall shape and magnitude. We refer to the studies listed in Sec.~\ref{sc:RelatedWork}.
\subsection{Limitations and Future Work}
\label{sc:Limitations}
The remaining position errors, which are as low as $26.8 \si{\micro \metre}$, are significantly smaller than the manufacturer-provided repeatabilities of 60 \si{\micro \metre} for the KUKA robot and 15\si{\micro\metre} + 6\si[per-mode=symbol]{\micro\metre\per\metre} the Leica laser tracker. Thus, as explained in Sec.~\ref{sc:Repeatability}, we compare our calibration results not to the nominal repeatability but to the experimentally determined average distance of points to their cluster center from the repeatability experiment, which we have found to be 6.26 \si{\micro \metre}. Consequently, we conclude that some unmodeled, deterministic effects remain. Drifts in the laser tracker measurements may account for the deviation between nominal and estimated repeatability and, in turn, for the discrepancy between the observed calibration error and the estimated repeatability. Some further effects might be considered as sources for the remaining error. While modeling compliant behavior using one-dimensional, linear springs is a standard approach for articulated robots, limitations of these assumptions may still account for some remaining error. A higher-resolution or coupled model of the joints' behavior would likely further improve accuracy, but is limited by the amount of data available. Similarly, the linear elongation model for thermally induced geometry changes may be inherently insufficient to capture some thermal effects. This may include twisting as well as higher-order relationships or time-dependent ones. 
Despite the strong performance within our tested data sets, the model’s generalization to unseen conditions may be limited in certain cases. Firstly, we restrict our analysis to thermal steady states with influences only due to the environment, such that generalization to unseen thermal states may be poor. For more generally applicable thermal compensation, a stronger thermal model could become necessary. Secondly, no backlash compensation is included, which limits accuracy whenever a unidirectional approach to a pose cannot be guaranteed. These areas represent opportunities for future research.
\section{Conclusion}
\label{sc:Conclusion}
In applications where we cannot solely rely on feedback control to guarantee end-effector accuracy, an accurate forward model is essential. This study presents a unified calibration framework that jointly identifies geometric offsets, compliance, gear transmission errors, and thermal deformation from purely static, end-effector measurements.
The core novelty of our work is the integrated and detailed identification of all these effects in an interpretable manner. A high-resolution model of joint and gear transmission errors, as proposed here, identified solely from end-effector measurements, has not been previously investigated. We validate our method extensively to ensure consistent performance, robust identifiability, and stable behavior when reduced training data is available. This allows us to confidently report a very high accuracy with a remaining mean position error as low as $26.8 \si{\micro \metre}$ and a maximum position error of $97.4 \si{\micro \metre}$ as verified under a rigorous cross-validation scheme. Comparisons to the setup's experimentally determined repeatability are discussed.
The approach of replacing joint transformations in the kinematic chain with their augmented version is inherently transferable to other open-chain robot designs. It also directly enables the use of derived model versions for settings where calibration is desired only for certain effects or in certain joints.

\ifCLASSOPTIONcaptionsoff
  \newpage
\fi



\bibliographystyle{IEEEtran}
\bibliography{bibliography.bib}
\end{document}